# Approximation Algorithms for *K*-Modes Clustering

Zengyou He

Department of Computer Science and Engineering, Harbin Institute of Technology,

92 West Dazhi Street, P.O Box 315, Harbin 150001, P. R. China

zengyouhe@yahoo.com

**Abstract** In this paper, we study clustering with respect to the *k*-modes objective function, a natural formulation of clustering for categorical data. One of the main contributions of this paper is to establish the connection between *k*-modes and *k*-median, i.e., the optimum of *k*-median is at most twice the optimum of *k*-modes for the same categorical data clustering problem. Based on this observation, we derive a deterministic algorithm that achieves an approximation factor of 2. Furthermore, we prove that the distance measure in *k*-modes defines a metric. Hence, we are able to extend existing approximation algorithms for metric *k*-median to *k*-modes. Empirical results verify the superiority of our method.

## 1 Introduction

The *k*-modes algorithm [1] extends the *k*-means paradigm to cluster categorical data by using (1) a simple matching dissimilarity measure for categorical objects, (2) modes instead of means for clusters, and (3) a frequency-based method to update modes in the k-means fashion to minimize the clustering cost function of clustering. Because the *k*-modes algorithm uses the same clustering process as *k*-means, it preserves the efficiency of the *k*-means algorithm.

Although the *k*-modes algorithm is very efficient, it suffers two well-known problems as *k*-means algorithm: the solutions are only locally optimal and their qualities are sensitive to the initial conditions. Hence, approximation algorithms with performance guarantees are desired for *k*-modes clustering. However, such kinds of approximation algorithms are not available so far.

In this paper, we study clustering with respect to the *k*-modes objective function. We first establish the connection between *k*-modes problem and well-known *k*-median problem by proving that the optimum of *k*-median is at most the twice the optimum of *k*-modes for the same categorical data clustering problem. Based on this observation, we derive a deterministic algorithm that achieves an approximation factor of 2. Furthermore, we prove that the distance measure in *k*-modes defines a metric. Hence, we are able to extend existing approximation algorithms for metric *k*-median [e.g., 2-4] to *k*-modes.

## 2 K-Modes Clustering

Let *X, Y* be two categorical objects described by *m* categorical attributes. The simple dissimilarity measure between *X* and *Y* is defined by the total mismatches of the corresponding attribute values of the two objects. The smaller the number of mismatches is, the more similar the two objects. Formally,

$$d(X,Y) = \sum_{j=1}^{m} \delta(x_j, y_j) \qquad (1)$$

where

$$\delta(x_j, y_j) = \begin{cases} 0 & (x_j = y_j) \\ 1 & (x_j \neq y_j) \end{cases} \qquad (2)$$

Let $S$ be a set of categorical objects described by $m$ categorical attributes $A_1, ..., A_m$. A mode of $S = \{X_1, X_2, ..., X_n\}$ is a vector $Q = [q_1, q_2, ..., q_m]$ that minimizes

$$D(S,Q) = \sum_{i=1}^{n} d(X_i, Q) \qquad (3)$$

Here, $Q$ is not necessarily an object of $S$.

Let $n_{c_{k,r}}$ be the number of objects having the $k$th category $c_{k,r}$ in attribute $A_r$ and $f(A_r = c_{k,r}) = \frac{n_{c_{k,r}}}{n}$ the relative frequency of category $n_{c_{k,r}}$ in $S$. The function $D(S,Q)$ is minimized iff $f(A_r = q_r) \geq f(A_r = c_{k,r})$ for $q_r \neq c_{k,r}$ and all $r = 1, ..., m$.

The optimization problem for partitioning a set of $n$ objects described by $m$ categorical attributes into $k$ clusters $S_1, S_2, ...S_k$ becomes

$$\text{Minimize } \sum_{i=1}^{k} \sum_{X \in S_i} d(X, Q_i) \qquad (4)$$

where $Q_i$ is the mode of cluster $S_i$.

In the above $k$-modes clustering problem, the representative point of each cluster $S_i$, i.e., the mode $Q_i$, is not necessarily contained in $S_i$. If we restrict the representative point to be in $S_i$, it becomes well-known $k$-median problem. In the next section, we will show that the optimum of $k$-median is at most the twice optimum of $k$-modes for the same categorical data clustering problem. Hence, the loss is modest even restricting representatives to points contained in original set.

## 3 Approximation Algorithms

**Lemma 1:** Let $S$ be a set of $n$ categorical objects described by $m$ categorical attributes and $Q$ be the mode of $S$. Then, there exists a $X_i \in S$ such that

$$\sum_{j=1}^{n} d(X_j, X_i) \leq 2 \sum_{j=1}^{n} d(X_j, Q) \qquad (5)$$

**Proof:** To prove the lemma, we only need to show the following inequality holds.

$$\sum_{i=1}^{n} \sum_{j=1}^{n} d(X_j, X_i) \leq 2n \sum_{j=1}^{n} d(X_j, Q) \qquad (6)$$

Recalling that $n_{c_{k,r}}$ is the number of objects having the *k*th category $c_{k,r}$ in attribute $A_r$ and $f(A_r = c_{k,r}) = \frac{n_{c_{k,r}}}{n}$ is the relative frequency of category $n_{c_{k,r}}$ in **S**. Without loss of generality, assume that the $q_r = c_{1,r}$ in attribute $A_r$, i.e., the first category $c_{1,r}$ in attribute $A_r$ has largest relative frequency in **S** for all $r = 1, …, m$.

Considering the left-hand side of (6), the contribution of the *r*th attribute is $n^2 \sum_k f(A_r = c_{k,r})(1 - f(A_r = c_{k,r}))$; On the other hand, the contribution of the *r*th attribute to the right-hand side is $n^2(1 - f(A_r = c_{1,r}))$. Hence, we have

$$\frac{n^2 \sum_k f(A_r = c_{k,r})(1 - f(A_r = c_{k,r}))}{n^2(1 - f(A_r = c_{1,r}))} = \frac{\sum_k f(A_r = c_{k,r})(1 - f(A_r = c_{k,r}))}{1 - f(A_r = c_{1,r})}$$

$$= \frac{\sum_k f(A_r = c_{k,r}) - \sum_k f^2(A_r = c_{k,r})}{1 - f(A_r = c_{1,r})}$$

$$= \frac{1 - \sum_k f^2(A_r = c_{k,r})}{1 - f(A_r = c_{1,r})} = \frac{1 - f^2(A_r = c_{1,r}) - \sum_{k \neq 1} f^2(A_r = c_{k,r})}{1 - f(A_r = c_{1,r})}$$

$$= 1 + f(A_r = c_{1,r}) - \frac{\sum_{k \neq 1} f^2(A_r = c_{k,r})}{1 - f(A_r = c_{1,r})} \leq 1 + f(A_r = c_{1,r}) \leq 2$$

Summing over *r*, we can verify inequality (6). □

**Lemma 2:** The optimum of *k*-median is at most the twice the optimum of *k*-modes for the same categorical data clustering problem.

**Proof:** Let $OPT^O = \{(S_1^O, Q_1), (S_2^O, Q_2), …, (S_k^O, Q_k)\}$ be the optimal solution of *k*-modes clustering, and $OPT^E = \{(S_1^E, Y_1), (S_2^E, Y_2), …, (S_k^E, Y_k)\}$ be the optimal solution of *k*-median clustering. According to Lemma 1, we can find a solution $P^E = \{(S_1^O, W_1), (S_2^O, W_2), …, (S_k^O, W_k)\}$ for *k*-median clustering such that $\sum_{X \in S_i^O} d(X, W_i) \leq 2 \sum_{X \in S_i^O} d(X, Q_i)$ for all $i = 1, …, k$. When considering all partitions, we have $\sum_{i=1}^k \sum_{X \in S_i^O} d(X, W_i) \leq 2 \sum_{i=1}^k \sum_{X \in S_i^O} d(X, Q_i)$. Furthermore, since $OPT^E$ is the optimal solution of *k*-median clustering, we have $\sum_{i=1}^k \sum_{X \in S_i^E} d(X, Y_i) \leq \sum_{i=1}^k \sum_{X \in S_i^O} d(X, W_i)$. Thus, we obtain $\sum_{i=1}^k \sum_{X \in S_i^E} d(X, Y_i) \leq 2 \sum_{i=1}^k \sum_{X \in S_i^O} d(X, Q_i)$, i.e., the optimum of *k*-median is at most the twice the optimum of *k*-modes for the same categorical data clustering

problem. □

By enumerating all *k*-subsets of ***S***, we could find the optimal solution of *k*-median problem. Therefore, by Lemma 2, we can solve the *k*-modes clustering deterministically in time $O(kn^{k+1})$, for a 2-approximation to the optimal solution. That is, we can derive a deterministic algorithm that achieves an approximation factor of 2.

The above deterministic algorithm is feasible for small *k*; for larger *k*, we can use existing efficient approximation algorithms for metric *k*-median [e.g., 2-4] since the distance measure in *k*-modes defines a metric (as shown in Lemma 3). That is, the distance is nonnegative, symmetric, satisfy the triangle inequality, and the distance between points *X* and *Y* is zero if and only if $X = Y$.

**Lemma 3**: The distance measure *d* in *k*-modes is a valid distance metric, such that: (1) $d(X,Y) > 0, \forall X \neq Y$, (2) $d(X,Y) = 0, \forall X = Y$, (3) $d(X,Y) = d(Y,X)$ and (4) $d(X,Y) + d(Y,Z) \geq d(X,Z), \forall X, Y, Z$

**Proof.** It is easy to verify that the first three properties are true. We prove the fourth statement is true. To simplify the presentation, we view the points *X*, *Y* and *Z* as sets of elements. Thus, $d(X,Y) = m - |X \cap Y|$, $d(Y,Z) = m - |Y \cap Z|$, $d(X,Z) = m - |X \cap Z|$. To prove the lemma, we only need to show the following inequality holds:

$2m - |X \cap Y| - |Y \cap Z| \geq m - |X \cap Z|$, i.e., $|X \cap Y| - |X \cap Z| + |Y \cap Z| \leq m$

We have $|X \cap Y| - |X \cap Z| + |Y \cap Z| \leq |X \cap (Y - Z)| + |Y \cap Z|$

$\leq |Y - Z| + |Y \cap Z| = |Y| = m$

Thus, the fourth statement is true. □

By Lemma 3, we know that the *k*-median problem with distance measure *d* in *k*-modes is a metric *k*-median problem. Therefore, one obvious idea for designing approximation algorithms for *k*-modes clustering is the direct adaptation and use of existing approximation algorithms for metric *k*-median [e.g., 2-4]. More precisely, as shown in Theorem 1, one α-approximation algorithm for metric *k*-median is a 2α-approximation algorithm for *k*-modes clustering by Lemma 2 and Lemma 3.

**Theorem 1**: For the same categorical data clustering problem with respect to the *k*-modes objective function, any α-approximation algorithm for metric *k*-median will be a 2α-approximation algorithm for *k*-modes clustering, where α is the approximation ratio guaranteed.

**Proof.** Trivial. □

According to Theorem 1, we are able to select effective α-approximation algorithm in metric *k*-median literature [e.g., 2-4] to approximate *k*-modes clustering with approximation ratio 2α is guaranteed.

# 4 Empirical Studies

A performance study has been conducted to evaluate our method. In this section, we describe those experiments and the results. We ran our deterministic 2-approximation algorithm on real-life datasets obtained from the UCI Machine Learning Repository [5] to test its clustering performance against original *k*-modes algorithm.

### 4.1 Real Life Datasets and Evaluation Method

We experimented with two real-life datasets: the Congressional Votes dataset and Mushroom dataset, which were obtained from the UCI Machine Learning Repository [5]. Now we will give a brief introduction about these datasets.

- ✓ **Congressional Votes:** It is the United States Congressional Voting Records in 1984. Each record represents one Congressman's votes on 16 issues. All attributes are Boolean with Yes (denoted as *y*) and No (denoted as *n*) values. A classification label of Republican or Democrat is provided with each record. The dataset contains 435 records with 168 Republicans and 267 Democrats.
- ✓ **The Mushroom Dataset:** It has 22 attributes and 8124 records. Each record represents physical characteristics of a single mushroom. A classification label of poisonous or edible is provided with each record. The numbers of edible and poisonous mushrooms in the dataset are 4208 and 3916, respectively.

Validating clustering results is a non-trivial task. In the presence of true labels, as in the case of the data sets we used, the clustering accuracy for measuring the clustering results was computed as follows. Given the final number of clusters, *k*, clustering accuracy *r* was defined as: $r = \frac{\sum_{i=1}^{k} a_i}{n}$, where *n* is the number of records in the dataset, $a_i$ is the number of instances occurring in both cluster *i* and its corresponding class, which had the maximal value. In other words, $a_i$ is the number of records with the class label that dominates cluster *i*. Consequently, the clustering error is defined as $e = 1 - r$.

Furthermore, we also compare the objective function values produced by both algorithms since such measure is non-subjective and provides hints on goodness of approximation.

### 4.2 Experimental Results

We studied the clusterings found by our algorithm and the original *k*-modes algorithm [1]. For the *k*-modes algorithm, we use the first *k* distinct records from the data set to construct initial *k* modes. That is, we use one run to get the clustering outputs for *k*-modes.

On the congressional voting dataset, we let the algorithms produce two clusters, i.e., *k*=2. Table 1 shows the clusters, class distribution, clustering errors and objective function values produced by two algorithms. As Table 1 shows, *k*-modes algorithm and our algorithm have similar performance. In particular, the objective function value of our algorithm is only a little lower than that of *k*-modes. It provides us hints that local heuristics based *k*-modes algorithm could find good solutions in some cases. However, as shown in the next experiment, such algorithm has poor worst-case performance guarantee and can produce very poor clustering output.

**Table 1:** Clustering results on votes data

| k-modes (Clustering Error: 0.136, Objective Function Value: 1706) | | |
|---|---|---|
| Cluster NO | No of Republicans | No of Democrats |
| 1 | 154 | 45 |
| 2 | 14 | 222 |
| Our 2-Approximation Algorithm (Clustering Error: 0.149, Objective Function Value: 1701) | | |
| Cluster NO | No of Democrats | No of Democrats |
| 1 | 158 | 55 |
| 2 | 10 | 212 |

Table 2 contrasts the clustering results on mushroom dataset. The number of clusters is still set to be 2 since there are two natural clusters in this dataset. As shown in Table 2, our algorithm performs much better than *k*-modes algorithm with respect to clustering accuracy and objective function values. It further empirically confirms the fact that our algorithm deserves good performance guarantee.

**Table 2:** Clustering results on mushroom data

| k-modes (Clustering Error: 0.435, Objective Function Value: 63015) | | |
|---|---|---|
| Cluster NO | No of Edible | No of Poisonous |
| 1 | 1470 | 1856 |
| 2 | 2738 | 2060 |
| Our 2-Approximation Algorithm (Clustering Error: 0.121, Objective Function Value: 62512) | | |
| Cluster NO | No of Edible | No of Poisonous |
| 1 | 4182 | 960 |
| 2 | 26 | 2956 |

# 5 Literature Review

Clustering categorical data is an important research topic in data mining. Recently, many algorithms have been proposed in the literature. In this paper, our focus is *k*-modes type clustering for categorical data. Hence, we will review only those *k*-modes related papers.

K-modes, an algorithm extending the *k*-means paradigm to categorical domain is introduced in [1]. New dissimilarity measures to deal with categorical data is conducted to replace means with modes, and a frequency based method is used to update modes in the clustering process to minimize the clustering cost function. Based on *k*-modes algorithm, [5] proposes an adapted mixture model for categorical data, which gives a probabilistic interpretation of the criterion optimized by the *k*-modes algorithm. A fuzzy *k*-modes algorithm is presented in [6] and tabu search technique is applied in [7] to improve fuzzy *k*-modes algorithm. An iterative initial-points refinement algorithm for categorical data is presented in [8]. A genetic clustering algorithm (called GKMODE) by integrating a *k*-modes algorithm is given in [9].

As can be seen from above literature review, to overcome locally optimal in *k*-modes clustering, some techniques such as tabu search and genetic algorithm have been investigated to find globally optimal solution. However, they cannot provide approximation guarantees. Thus, effective approximation algorithms should be designed for *k*-modes clustering. To the best of our knowledge, such kinds of approximation algorithms are still not available to date.

# 6 Conclusions

This paper reveals an interesting fact that the optimum of $k$-median is at most the twice the optimum of $k$-modes for the same categorical data clustering problem. This observation makes possible the study of $k$-modes clustering problem from a metric $k$-median perspective. From this viewpoint, effective approximation algorithms are designed and empirically studied.